\documentclass{article}

\usepackage[preprint]{style}
\usepackage{graphicx}

\usepackage[utf8]{inputenc} 
\usepackage[T1]{fontenc}    
\usepackage{hyperref}       
\usepackage{url}            
\usepackage{booktabs}       
\usepackage{amsfonts}       
\usepackage{nicefrac}       
\usepackage{microtype}      
\usepackage{xcolor}         

\usepackage{caption}
\usepackage{subcaption}

\title{Tech Report: \\ One-stage Lightweight Object Detectors}

\author{%
  Deokki Hong\thanks{Work done during internship in CLOVA ImageVision, NAVER}\\  
  College of Computing, Yonsei University \\
  \texttt{dk.hong@yonsei.ac.kr} \\
}

\begin{document}

\maketitle

\begin{abstract}
  This work is for designing one-stage lightweight detectors which perform well in terms of mAP and latency.
  With baseline models each of which targets on GPU and CPU respectively, various operations are applied instead of the main operations in backbone networks of baseline models.
  In addition to experiments about backbone networks and operations, several feature pyramid network (FPN) architectures are investigated.
  Benchmarks and proposed detectors are analyzed in terms of the number of parameters, Gflops, GPU latency, CPU latency and mAP, on MS COCO dataset which is a benchmark dataset in object detection.
  This work propose similar or better network architectures considering the trade-off between accuracy and latency.
  For example, our proposed GPU-target backbone network outperforms that of \emph{YOLOX}-tiny which is selected as the benchmark by 1.43x in speed and 0.5 mAP in accuracy on \emph{NVIDIA GeForce RTX 2080 Ti} GPU.
\end{abstract}
\section{Introduction}
\label{intro}

Object detection~\cite{rcnn, fastrcnn} is one of the various vision tasks that localizes and classifies objects in a scene.
In recent years, object detection is applied to numerous fields such as unmanned stores, and face-recognition based security systems.
Early Studies about object detection are based on two-stage detectors~\cite{fasterrcnn}, which show high performance but low hardware efficiency.
Nowadays, the more object detection permeates in real life, the higher demand increases for lightweight detectors.
For instance, there may need real-time object detection in a field of surveillance systems, or some constraints such as battery limit and computing power can exist in edge devices.
However, the efficiency of a detector is affected by not only operations in the network, but also hardware architectures that the detector is executed on.
For instance, inverted residual bottleneck, which is proposed in MobileNetv2~\cite{mobilenetv2}, is designed for better efficiency while barely sacrificing accuracy.
Meanwhile, Google's TPU~\cite{tpu} is one of the most remarkable hardware which is optimized for executing DNNs.
Unfortunately, inverted residual bottleneck performs poorly on TPU for its architectural advantage is not suitable for exploiting TPU~\cite{edgetpunas}.
Therefore, it is necessary to design networks while considering properties of operations and hardware architectural features comprehensively.

In this work, we examined novel one-stage lightweight detectors~\cite{yolox2021, picodet} and various modern operations in terms of accuracy and latency.
Based on the examination, we propose the best operations and architectures on GPU and CPU, respectively.
In GPU experiments, our suggested object detector is based on \emph{YOLOX}~\cite{yolox2021} and it adopts fused inverted residual botteneck~\cite{tan2021efficientnetv2} in the front and inverted residual bottleneck in the back.
It outperforms YOLOX-tiny 1.43x in speed and 0.5 mAP in accuracy.
In CPU experiments, despite that YOLOX-tiny is the best in terms of mAP, our suggested one which is based on PP-PicoDet~\cite{picodet} shows only 74\% of the number of parameters and 1.12x in speed while sacrificing 1.3 mAP in accuracy.

\section{Related works}
\label{related_works}

\subsection{One-stage detector}
\label{onestage}

One-stage detector often consists of a backbone network, feature pyramid network (FPN)~\cite{fpn}, and detection head.
Backbone network is a general feed-forward network and it extracts meaningful properties from input images.
Unlike two-stage detectors, one-stage detectors directly use features from backbone networks in bounding box regression and classification.
Thus, backbone network is highly related to the performance of the detector.

From each stage of a backbone network, various sizes of features are obtained.
High-resolution features from earlier stages have weak semantics, while low-resolution features from later stages have strong semantics.
FPN combines these various sizes of features and makes features from earlier stages have rich information.

Detection head is the most important part of detectors.
The two tasks of object detection, bounding box regression and classification, are done in detection head by using features from backbone network and FPN.
Detection head is highly related to the loss function of the detector.

There are two baselines in this work, each of which targets GPU and CPU, respectively.
YOLOX~\cite{yolox2021} is one of the latest one-stage object detectors, which is used as the main baseline for GPU in this work.
The backbone network of YOLOX is CSPDarknet and the FPN of YOLOX is path aggregation feature pyramid network (PAFPN)~\cite{pafpn}.
PP-PicoDet~\cite{picodet} is also a novel one-stage detector that targets on CPU.
The backbone network of PP-PicoDet is PP-LCNet (lightweight CPU convolutional neural network)~\cite{lcnet}.
The FPN of PP-PicoDet is Lightweight CPU Path Aggregation Network (LCPAN), which is a variant of PAFPN.

\subsection{Operations}
\label{operations}
There are several works that pursue efficiency while trying to barely sacrifice performance.
CSP layer~\cite{wang2020cspnet}, which is the main operation in YOLOX backbone network, is an architecture for achieving abundant gradient combinations while reducing computational costs.
Depthwise separable convolution operation, which is the main operation of PP-LCNet, was proposed in MobileNet~\cite{mobilenet} which is one of the early works in this field.
It reduces computational costs by using depthwise convolution operation and combines information among channels with a $1 \times 1$ convolution operation.
MobileNetv2~\cite{mobilenetv2} is a follow-up research of MobileNet and it proposed inverted residual bottleneck architecture, which is a dominant architecture in mobile settings.
Based on depthwise separable convolution, it expands input channels of the depthwise convolution operation with a $1 \times 1$ convolution operation in front of that.
EfficientNetv2~\cite{tan2021efficientnetv2} tackled MobileNetv2 that using inverted residual bottleneck in the whole network is not efficient, and it proposed fused inverted residual bottleneck architecture, which replaces the first $1 \times 1$ convolution and the depthwise convolution operation of inverted residual bottleneck with a convolution operation, for replacing inverted residual bottleneck at the front of networks.
X block used in RegNet~\cite{regnet} is a result of designing generalized architecture across settings, and sandglass bottleneck~\cite{sandglass} is designed to improve inverted residual bottleneck for mitigating information loss and gradient confusion.
Blueprint separable convolution~\cite{bsconv} is a variant of depthwise separable convolution which aims for improving MobileNet by replacing cross-kernel correlations in depthwise separable convolution with intra-kernel correlations.

\subsection{Feature pyramid network}
\label{fpn}
In a general feed-forward neural network, high-resolution features from early layers have weak semantics while low-resolution features from later layers have strong semantics.
FPN is an architecture for combining various sizes of features to make all features have strong semantics.
The FPN architectures in both YOLOX and PP-PicoDet are based on PAFPN.
PAFPN is based on FPN architecture which adds bottom-up path augmentation for enhancing semantics of features by propagating low-level features to others.
SepFPN is an FPN architecture based on PAFPN, which removes the bottom-up path in PAFPN and adds residual paths from inputs to outputs of FPN.
It pursues faster latency even at the expense of accuracy a little.

\section{Lightweight detector design}
\label{design}

\subsection{Micro architectures for backbone network}
In GPU experiments, CSP layer in CSPDarknet is replaced with several bottleneck architectures, such as inverted residual bottleneck (MBConv), fused inverted residual bottleneck, RegNet bottleneck, and sandglass bottleneck to verify whether CSP layer is the best architecture in YOLOX backbone network.
EfficientNetv2~\cite{tan2021efficientnetv2} proposed that using fused inverted residual bottleneck to the front of networks and inverted residual bottleneck to the rest is good for both accuracy and efficiency.
In this paper, each network that comprises inverted residual bottleneck only, fused inverted residual bottleneck only, and use both operations simultaneously are addressed.
A policy that uses inverted residual bottleneck and fused inverted residual bottleneck in a single network is named mixed inverted residual bottleneck.

In CPU experiments, depthwise separable convolution operation and blueprint separable convolution operation are examined with PP-PicoDet.
For a fair comparison with YOLOX baseline, channels of each block are set the same as those of YOLOX.
Additionally, FPN architecture and detection head in PP-PicoDet are replaced with those of YOLOX.


\subsection{Feature pyramid network}
The main operation of YOLOX's PAFPN is CSP layer and that of LCPAN is depthwise separable convolution, respectively.
In addition, the major difference between these two FPN is whether channels of input features are equalized before FPN operation or not.
In PAFPN, channels of input features are not equalized.
Rather, channels of FPN outputs are equalized before they are fed to the detection head.
It leads to better performance in terms of accuracy but is bad for latency because channels in FPN are large.
On contrary, in LCPAN, channels of input features are equalized in front of FPN.
Then, channels of output features are the same, while channels in FPN are reduced.
SepFPN is based on PAFPN of YOLOX.
That is, the main operation of SepFPN is CSP layer and channels of input features are not equalized.
In this work, a modified PAFPN architecture is proposed, which replaces concatenation operations in FPN with sum.
By doing so, channels in FPN can be reduced while rich semantics in feature maps are expected to preserve.
This technique is applied to both YOLOX's PAFPN and PP-PicoDet's LCPAN and examined.
\section{Experimental results}
\label{results}

\subsection{Experimental settings}
\label{setting}
For a fair comparison, a detection head from YOLOX is applied in both baselines.
That is, several operations, backbones, and FPN architectures are targets in this work.
In GPU experiments, expand ratios of bottleneck architectures except sandglass bottleneck are set to 1, and that of sandglass bottleneck is set to 0.5.
Hyperparameters related to network design (e.g., the number of blocks, channels of each block, etc.) are set as the same with YOLOX-tiny.
While training networks, any other hyperparameters except network architectures follow the default settings of YOLOX.
In GPU experiments, \emph{NVIDIA GeForce RTX 2080 Ti} is used to measure GPU latency.
In CPU experiments, \emph{Intel(R) Core(TM) i9-9900K CPU @ 3.60GHz} is used to measure CPU latency.
While measuring latency, mini batch size and the number of threads are set to 1.

\subsection{Baseline latency breakdown}
\label{sec:latency}

\begin{figure}[t]
  \centering
    \includegraphics[width=\columnwidth]{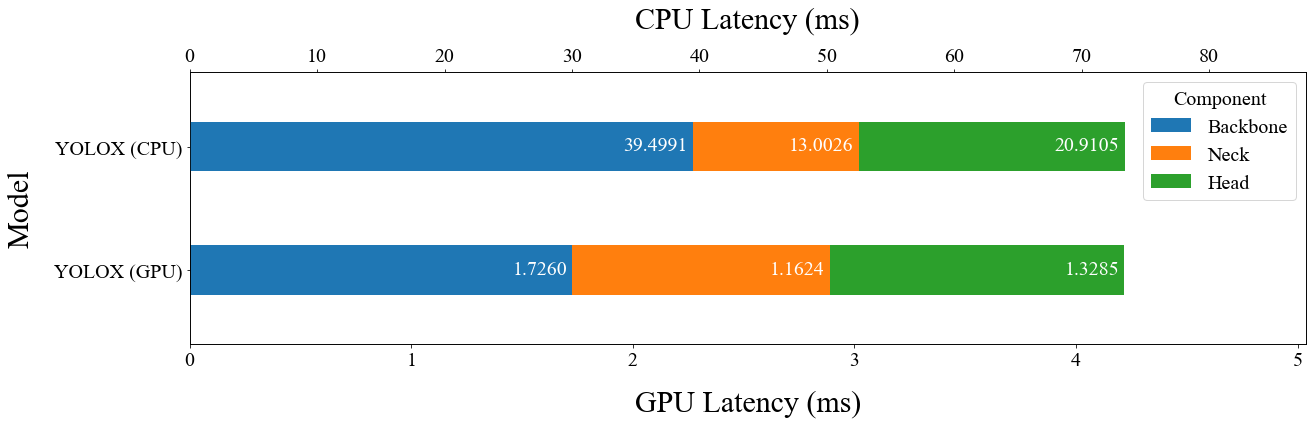}
  \caption{Latency breakdown of baseline model.}
  \label{fig:breakdown}
\end{figure}

\figureautorefname~\ref{fig:breakdown} shows GPU and CPU latency of the baseline model. 
The backbone network accounts for 40\% of the total latency on GPU, and 53\% on CPU.
Thus, it is critical to reducing backbone latency to lighten detectors.
FPN, which is also the target to improve in this work, accounts for 27\% of total latency on GPU and 18\% on CPU.
Detection head occupies more compared to FPN.
However, because it is highly related to the loss function of detectors, detection head is fixed in all experiments for a fair comparison.

\subsection{GPU-target detector}
\label{sec:gpu}


\begin{figure}[b]
  \centering
    \begin{subfigure}[b]{0.497\textwidth}
      \centering
      \includegraphics[width=\columnwidth]{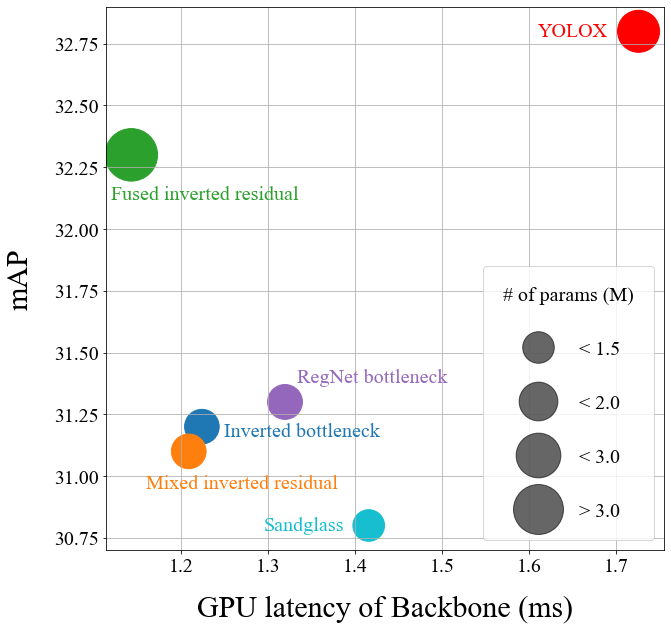}
      \caption{Backbone GPU Latency - mAP of CSPdarknet backbone and candidate operations.}
      \label{fig:gpu_backbone}
    \end{subfigure}
    \begin{subfigure}[b]{0.49\textwidth}
      \centering
      \includegraphics[width=\columnwidth]{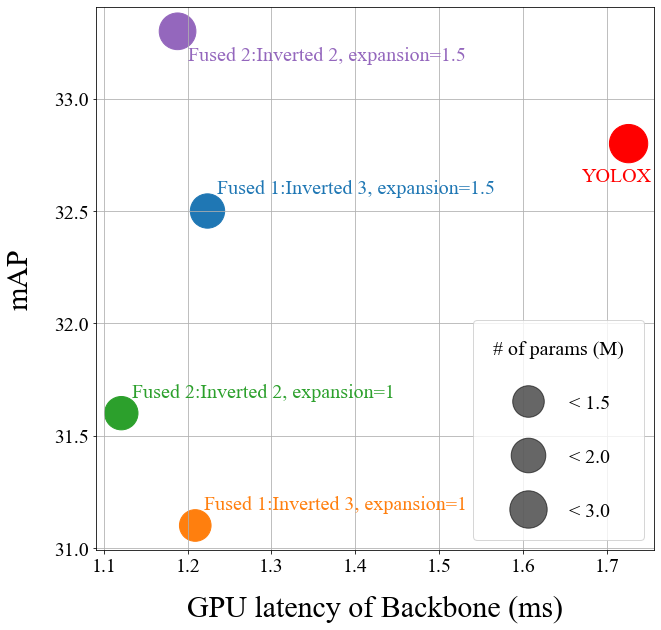}
      \caption{Ablation study for mixed inverted residual bottleneck.}
      \label{fig:gpu_ablation}
    \end{subfigure}
  \caption{Results of GPU experiments.}
  \label{fig:gpu}
  \vspace{-6mm}
\end{figure}

\figureautorefname~\ref{fig:gpu} shows GPU latency of backbone network and mAP depending on the main operation of backbone networks.
As shown in the \figureautorefname~\ref{fig:gpu_backbone}, YOLOX tiny is the best in terms of mAP, but its GPU latency is the worst.
A backbone network that consists of fused inverted residual bottleneck has the largest number of parameters.
However, its GPU latency is the fastest among all settings.
Fused inverted bottleneck is the only one that uses a $3 \times 3$ convolution operation rather than a $3 \times 3$ depthwise convolution operation.
$3 \times 3$ convolution operation is the most basic convolution operation, and it is highly optimized on GPU.
That's the reason a detector that adopts fused inverted residual bottleneck as the main operation of the backbone is the fastest on GPU even though the number of parameters is the largest.

This work focuses on the policy which is proposed in Efficientnetv2~\cite{tan2021efficientnetv2}; \emph{Use fused inverted residual bottleneck to the front of the network and inverted residual bottleneck to the rest}.
In \figureautorefname~\ref{fig:gpu_backbone}, mixed inverted residual bottleneck is noticeable because it is fast, lightweight, and has a lot of diversity in terms of design.
Mixed inverted residual bottleneck can leverage parallel computing and get better mAP by using fused inverted residual bottleneck while pursuing lightness by using inverted residual bottleneck.
Also, it has a lot of potential because the number of fused inverted residual bottleneck operations is an important design policy.
CSPDarknet, which is the backbone network of YOLOX, has 4 blocks.
Therefore, networks that use 1 or 2 fused inverted bottlenecks are examined.
Furthermore, because the number of parameters in mixed inverted bottleneck is smaller than that of YOLOX baseline, networks that set expand ratio to 1.5 is also investigated.
\figureautorefname~\ref{fig:gpu_ablation} shows the results of this ablation study.
Using 2 fused inverted residual bottlenecks and 2 inverted residual bottlenecks is better than 1 fused inverted residual bottleneck and 3 inverted residual bottlenecks, in terms of latency and mAP.
Moreover, a policy that uses a larger expand ratio is still faster than YOLOX baseline.
Therefore, the purple dot in \figureautorefname~\ref{fig:gpu_ablation} is chosen as the best detector in GPU experiments.



\subsection{CPU-target detector}

\begin{figure}[b]
  \centering
    \begin{subfigure}[b]{0.49\textwidth}
      \centering
      \includegraphics[width=\columnwidth]{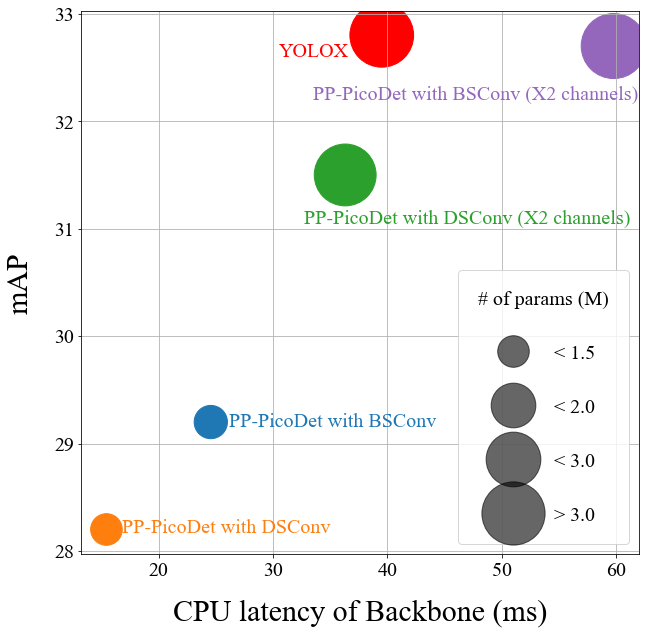}
      \caption{Backbone CPU Latency - mAP of PP-LCNet backbone and candidate operations.}
      \label{fig:cpu_backbone}
    \end{subfigure}
    \begin{subfigure}[b]{0.49\textwidth}
      \centering
      \includegraphics[width=\columnwidth]{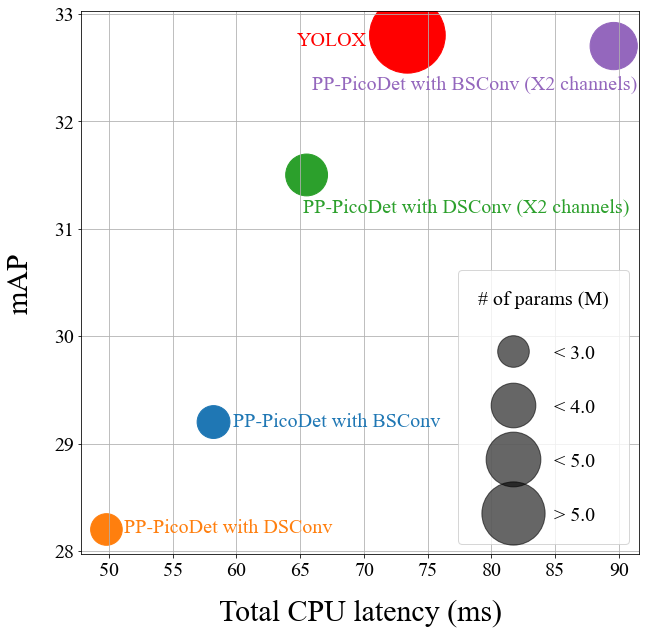}
      \caption{Total CPU Latency - mAP of PP-LCNet backbone and candidate operations.}
      \label{fig:cpu_total}
    \end{subfigure}
  \caption{Results of CPU experiments.}
  \label{fig:cpu}
\end{figure}

In this section, CSPDarknet, which is the backbone network of YOLOX, is replaced with PP-LCNet in PP-PicoDet.
In addition, blueprint separable convolution operation is applied to PP-LCNet as an alternative of depthwise separable convolution, which is the main operation of PP-LCNet.
The channels of PP-LCNet are set the same as those of CSPDarknet in order to compare fairly.
However, because PP-LCNet is much smaller than CSPDarknet, detectors that use twice as large as the default network are also examined.
Note that FPN architecture of these detectors is different from others in order to use the same detection head for a fair comparison.

\figureautorefname~\ref{fig:cpu_backbone} shows CPU latency of backbone network and mAP of the above detectors.
Using PP-LCNet as the backbone leads to generating lightweight detectors.
Detectors that adopt blueprint separable convolution operation as the main operation are slower and show higher mAP compared to those which adopt depthwise separable convolution operation.
Unfortunately, detectors that adopt PP-LCNet as the backbone network show low mAP as much as small the number of parameters they have compared to that of YOLOX baseline.
To improve mAP, backbone networks with larger channels are also analyzed.
They show a lot of improvement in mAP, while sacrificing latency.
Even though the number of parameters in each backbone is similar to that of YOLOX, either latency or mAP is worse than that of YOLOX.
When the number of channels in the backbone network is changed, that in FPN also should be changed.
However, because detection head is fixed in all experiments, the output channels of PP-LCNet with larger channels are equalized and therefore their FPN architectures are smaller than others.
\figureautorefname~\ref{fig:cpu_total} shows CPU latency of whole detector and mAP of the same detectors in \figureautorefname~\ref{fig:cpu_backbone}, for taking their FPN architectures into account.
A detector that adopts DSConv as the main operation and has larger channels shows faster latency and a much less number of parameters compared to YOLOX baseline.
The number of parameters is also a major constraint in real-world scenarios, and thus this setting can be a good solution when a fast detector with less number of parameters is needed.


\subsection{FPN architecture analysis}

\begin{figure}[t]
  \centering
    \begin{subfigure}[b]{0.49\textwidth}
      \centering
      \includegraphics[width=\columnwidth]{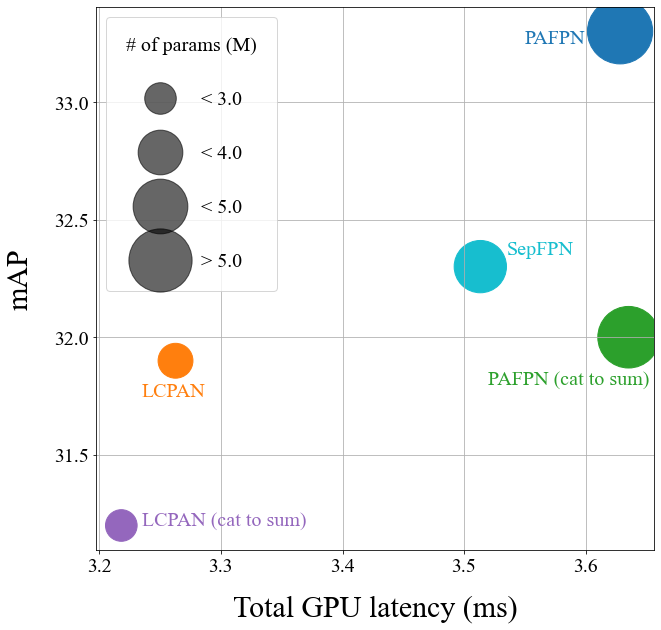}
      \caption{GPU experiments.}
      \label{fig:fpn_gpu}
    \end{subfigure}
    \begin{subfigure}[b]{0.49\textwidth}
      \centering
      \includegraphics[width=\columnwidth]{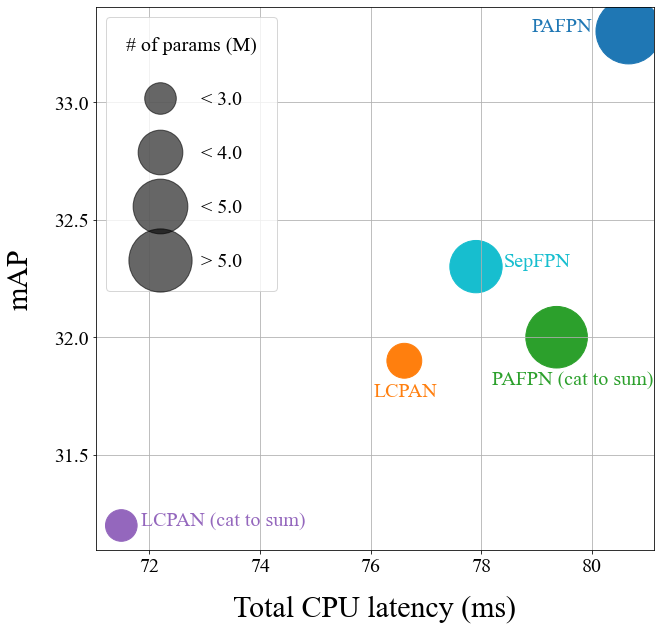}
      \caption{CPU experiments.}
      \label{fig:fpn_cpu}
    \end{subfigure}
  \caption{Results of FPN experiments.}
  \label{fig:fpn}
\end{figure}

In this section, the performances of FPN architectures are investigated with the backbone network which is searched in \sectionautorefname~\ref{sec:gpu}.
PAFPN, which is the default FPN architecture of YOLOX is the best in terms of mAP.
LCPAN, which is the default FPN architecture of PP-PicoDet is the best in terms of the number of parameters and latency.
Proposed FPN architectures (replace concatenation operation with sum) perform bad compare to baseline.
The expected effect of proposed FPN architectures is preserving semantics while reducing channels in FPN by summing features from different blocks.
However, it doesn't work as expected.
Especially, PAFPN-based proposed FPN architecture performs badly on GPU.
It means that reducing channels in FPN doesn't lead to reducing latency because GPU has huge parallel computing power.

\section{Conclusion}
\label{conclusion}

\begin{figure}[b!]
  \centering
  \vspace{-5mm}
    \begin{subfigure}[b]{0.49\textwidth}
      \centering
      \includegraphics[width=\columnwidth]{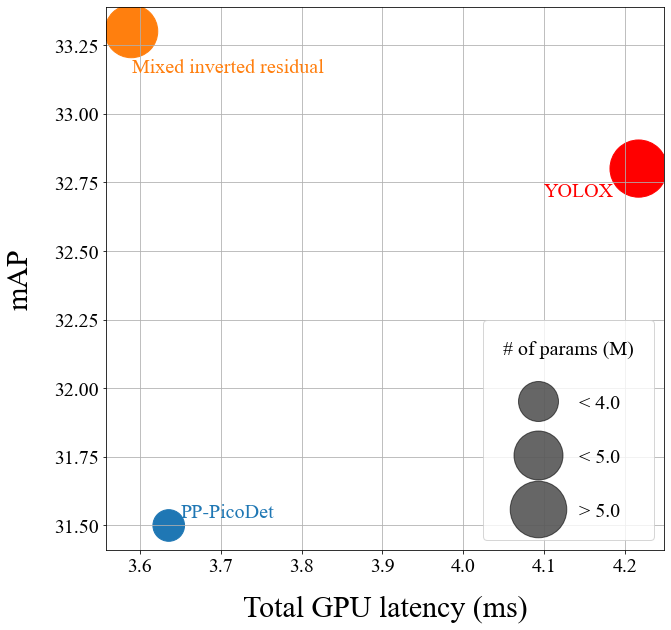}
      \caption{GPU experiments.}
      \label{fig:final_gpu}
    \end{subfigure}
    \begin{subfigure}[b]{0.49\textwidth}
      \centering
      \includegraphics[width=\columnwidth]{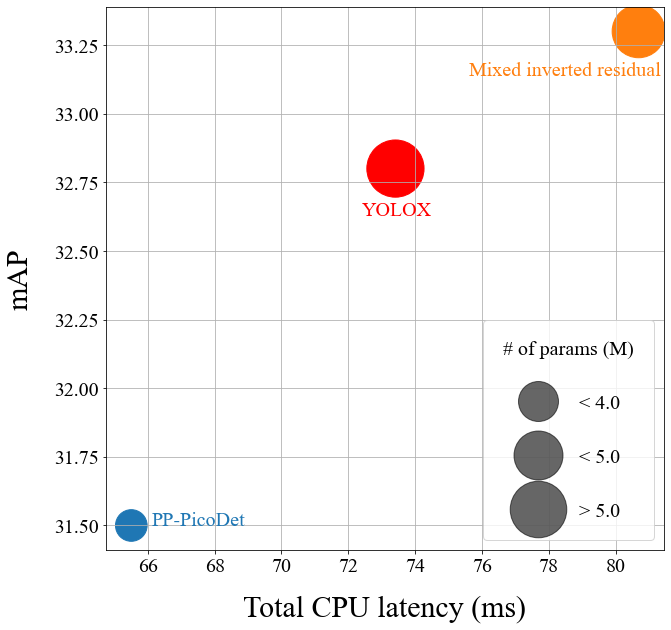}
      \caption{CPU experiments.}
      \label{fig:final_cpu}
    \end{subfigure}
  \caption{Results of searched networks.}
  \label{fig:final}
\end{figure}

This work analyzes the architectural design spaces of a novel one-stage detector.
Optimal architecture design in object detectors depends on target hardware and purpose.
\figureautorefname~\ref{fig:final} shows searched networks in \sectionautorefname~\ref{results} along with YOLOX baseline.
\figureautorefname~\ref{fig:final_gpu} denotes mAP and latency of the baseline, the GPU target detector, and the CPU target detector executed on GPU.
The GPU target detector which is marked with an orange dot shows the best mAP while achieving the best GPU latency, even though the CPU target detector has a less number of parameters.
Therefore, the GPU target detector searched in this paper is the best in both aspects.
\figureautorefname~\ref{fig:final_gpu} denotes mAP and latency of detectors the same as \figureautorefname~\ref{fig:final_gpu} executed on CPU.
The CPU target detector which is marked with a blue dot shows the worst mAP, yet it is much faster than others on CPU.
Further, it has a much less number of parameters compared to others.
Therefore, if there is not enough computing power and energy budget, the CPU target detector can be a good solution.
\figureautorefname~\ref{fig:final} doesn't plot FPN results. Nevertheless, FPN is also an important factor in detectors and should be designed carefully depending on the purpose.




{\small
\bibliographystyle{ieeetr}
\bibliography{refs}

\begin{thebibliography}{10}

\bibitem{rcnn}
R.~B. Girshick, J.~Donahue, T.~Darrell, and J.~Malik, ``Rich feature
  hierarchies for accurate object detection and semantic segmentation,'' in
  {\em {Proceedings of the IEEE/CVF Conference on Computer Vision and Pattern
  Recognition}}, pp.~580--587, 2014.

\bibitem{fastrcnn}
R.~B. Girshick, ``{Fast R-CNN},'' in {\em Proceedings of the IEEE International
  Conference on Computer Vision}, pp.~1440--1448, 2015.

\bibitem{fasterrcnn}
S.~Ren, K.~He, R.~B. Girshick, and J.~Sun, ``{Faster R-CNN: Towards Real-Time
  Object Detection with Region Proposal Networks},'' in {\em Advances in Neural
  Information Processing Systems}, 2015.

\bibitem{mobilenetv2}
M.~Sandler, A.~G. Howard, M.~Zhu, A.~Zhmoginov, and L.~Chen, ``{MobileNetV2:
  Inverted Residuals and Linear Bottlenecks},'' in {\em {Proceedings of the
  IEEE/CVF Conference on Computer Vision and Pattern Recognition}},
  pp.~4510--4520, 2018.

\bibitem{tpu}
N.~P. Jouppi, C.~Young, N.~Patil, D.~Patterson, G.~Agrawal, R.~Bajwa, S.~Bates,
  S.~Bhatia, N.~Boden, A.~Borchers, {\em et~al.}, ``{In-Datacenter Performance
  Analysis of a Tensor Processing Unit},'' in {\em ISCA}, 2017.

\bibitem{edgetpunas}
S.~Gupta and B.~Akin, ``{Accelerator-aware Neural Network Design using
  AutoML},'' {\em arXiv preprint arXiv:2003.02838}, 2020.

\bibitem{yolox2021}
Z.~Ge, S.~Liu, F.~Wang, Z.~Li, and J.~Sun, ``{YOLOX: Exceeding YOLO Series in
  2021},'' {\em arXiv preprint arXiv:2107.08430}, 2021.

\bibitem{picodet}
G.~Yu, Q.~Chang, W.~Lv, C.~Xu, C.~Cui, W.~Ji, Q.~Dang, K.~Deng, G.~Wang, Y.~Du,
  B.~Lai, Q.~Liu, X.~Hu, D.~Yu, and Y.~Ma, ``{PP-PicoDet: A Better Real-Time
  Object Detector on Mobile Devices},'' {\em arXiv preprint arXiv:2111.00902},
  2021.

\bibitem{tan2021efficientnetv2}
M.~Tan and Q.~Le, ``Efficientnetv2: Smaller models and faster training,'' in
  {\em International Conference on Machine Learning}, pp.~10096--10106, PMLR,
  2021.

\bibitem{fpn}
T.~Lin, P.~Doll{\'{a}}r, R.~B. Girshick, K.~He, B.~Hariharan, and S.~J.
  Belongie, ``{Feature Pyramid networks for Object Detection},'' in {\em
  {Proceedings of the IEEE/CVF Conference on Computer Vision and Pattern
  Recognition}}, pp.~2117--2125, 2017.

\bibitem{pafpn}
S.~Liu, L.~Qi, H.~Qin, J.~Shi, and J.~Jia, ``{Path Aggregation Network for
  Instance Segmentation},'' in {\em {Proceedings of the IEEE/CVF Conference on
  Computer Vision and Pattern Recognition}}, pp.~8759--8768, 2018.

\bibitem{lcnet}
C.~Cui, T.~Gao, S.~Wei, Y.~Du, R.~Guo, S.~Dong, B.~Lu, Y.~Zhou, X.~Lv, Q.~Liu,
  X.~Hu, D.~Yu, and Y.~Ma, ``{PP-LCNet: A Lightweight CPU Convolutional Neural
  Network},'' {\em arXiv preprint arXiv:2109.15099}, 2021.

\bibitem{wang2020cspnet}
C.-Y. Wang, H.-Y. Mark~Liao, Y.-H. Wu, P.-Y. Chen, J.-W. Hsieh, and I.-H. Yeh,
  ``{CSPNet: A new backbone that can enhance learning capability of CNN},'' in
  {\em Proceedings of the IEEE/CVF Conference on Computer Vision and Pattern
  Recognition Workshops}, pp.~390--391, 2020.

\bibitem{mobilenet}
A.~G. Howard, M.~Zhu, B.~Chen, D.~Kalenichenko, W.~Wang, T.~Weyand,
  M.~Andreetto, and H.~Adam, ``{MobileNets: Efficient Convolutional Neural
  Networks for Mobile Vision Applications},'' {\em arXiv preprint
  arXiv:1704.04861}, 2017.

\bibitem{regnet}
I.~Radosavovic, R.~P. Kosaraju, R.~B. Girshick, K.~He, and P.~Doll{\'{a}}r,
  ``{Designing Network Design Spaces},'' in {\em {Proceedings of the IEEE/CVF
  Conference on Computer Vision and Pattern Recognition}}, pp.~10428--10436,
  2020.

\bibitem{sandglass}
D.~Zhou, Q.~Hou, Y.~Chen, J.~Feng, and S.~Yan, ``{Rethinking Bottleneck
  Structure for Efficient Mobile Network Design},'' in {\em {European
  Conference on Computer Vision}}, pp.~680--697, 2020.

\bibitem{bsconv}
D.~Haase and M.~Amthor, ``{Rethinking Depthwise Separable Convolutions: How
  Intra-Kernel Correlations Lead to Improved MobileNets},'' in {\em Proceedings
  of the IEEE/CVF Conference on Computer Vision and Pattern Recognition},
  pp.~14600--14609, 2020.

\end{thebibliography}
}

\clearpage

\appendix
\section{Appendix}

\subsection{Experimental results with values}

The below tables show the stats of each detector searched in \sectionautorefname~\ref{results}.
A bold value represents the best value in each column.

\begin{table}[!h]
\centering
\begin{tabular}{|l||c|c|c|c|}
\hline
  Architecture
& \begin{tabular}[c]{@{}l@{}}backbone \# of\\ params (M)\end{tabular}
& \begin{tabular}[c]{@{}l@{}}backbone\\ Gflops\end{tabular}
& \begin{tabular}[c]{@{}l@{}}backbone GPU\\ latency (ms)\end{tabular}
& mAP
\\ \hline \hline

YOLOX tiny                 & 2.3723  &  2.6648  &  1.7260  & \textbf{32.8}
\\ \hline
Inverted bottleneck        & 1.6080  &  1.5884  &  1.2242  &  31.2
\\ \hline
Mixed inverted bottleneck  & 1.6167  &  1.6953  &  1.2091  &  31.1
\\ \hline
Fused inverted bottleneck  & 3.7042  & 3.9063   &  \textbf{1.1431}  & 32.3
\\ \hline
RegNet bottleneck          & 1.6329  & 1.6668   &  1.3197  & 31.3
\\ \hline
Sandglass bottleneck       & \textbf{1.3432}  & \textbf{1.3997}   & 1.4159   & 30.8
\\ \hline
\end{tabular}
\vspace{3mm}
\caption{Results of GPU experiments in \figureautorefname~\ref{fig:gpu_backbone}.}
\label{tab:gpu}
\vspace{-3mm}
\end{table}

\begin{table}[!h]
\centering
\begin{tabular}{|l||c|c|c|c|}
\hline
  Architecture
& \begin{tabular}[c]{@{}l@{}}backbone \# of\\ params (M)\end{tabular}
& \begin{tabular}[c]{@{}l@{}}backbone\\ Gflops\end{tabular}
& \begin{tabular}[c]{@{}l@{}}backbone GPU\\ latency (ms)\end{tabular}
& mAP
\\ \hline \hline

YOLOX tiny                 & 2.3723  &  2.6648  &  1.7260  & 32.8
\\ \hline
Fused 1: Inverted 3, expansion=1.0      &  \textbf{1.6167}  &  \textbf{1.6953}  &  1.2091  &  31.1
\\ \hline
Fused 2: Inverted 2, expansion=1.0      &  1.7978  &  2.5971  &  \textbf{1.1209}  &  31.6
\\ \hline
Fused 1: Inverted 3, expansion=1.5      &  1.9054  &  2.2505  &  1.2237  &  32.5
\\ \hline
Fused 2: Inverted 2, expansion=1.5      &  2.1771  &  3.6032  &  1.2046  &  \textbf{33.3}
\\ \hline
\end{tabular}
\vspace{3mm}
\caption{Results of GPU ablation studies in \figureautorefname~\ref{fig:gpu_ablation}.}
\label{tab:gpu_ablation}
\vspace{-3mm}
\end{table}

\begin{table}[!h]
\centering
\begin{tabular}{|l||c|c|c|c|}
\hline
  Architecture
& \begin{tabular}[c]{@{}l@{}}backbone \# of\\ params (M)\end{tabular}
& \begin{tabular}[c]{@{}l@{}}backbone\\ Gflops\end{tabular}
& \begin{tabular}[c]{@{}l@{}}backbone CPU\\ latency (ms)\end{tabular}
& mAP
\\ \hline \hline

YOLOX tiny              &   2.3723  &   2.6648  &   39.4991 &   \textbf{32.8}
\\ \hline
PP-PicoDet DSconv       &   \textbf{0.5845}  &   \textbf{0.6765}  &   \textbf{15.3962} &   28.2
\\ \hline
PP-PicoDet BSconv       &   0.6471  &   1.0118  &   24.5329 &   28.2
\\ \hline
PP-PicoDet DSconv ($\times$2 channels)       &   2.2358  &   2.3710  &   36.2975 &   31.5
\\ \hline
PP-PicoDet BSconv ($\times$2 channels)       &   2.4718  &   3.6565  &   59.8067 &   32.7
\\ \hline
\end{tabular}
\vspace{3mm}
\caption{Results of CPU experiments in \figureautorefname~\ref{fig:cpu_backbone}.}
\label{tab:cpu_backbone}
\vspace{-3mm}
\end{table}

\begin{table}[!h]
\centering
\begin{tabular}{|l||c|c|c|c|}
\hline
  Architecture & \# of params (M) & Gflops & CPU latency (ms) & mAP
\\ \hline \hline

YOLOX tiny                 & 5.0559  &  6.4510  &  73.4122  & \textbf{32.8}
\\ \hline
PP-PicoDet DSconv          & \textbf{3.2680}  &  \textbf{4.4628}  &  \textbf{49.8036}  &  28.2
\\ \hline
PP-PicoDet BSconv          & 3.3307  &  4.7980  &  58.0259  &  29.2
\\ \hline
PP-PicoDet DSconv ($\times$2 channels)  & 3.7591  &  5.5394   &  65.5060  &  31.5
\\ \hline
PP-PicoDet BSconv ($\times$2 channels)  & 3.9951  &  6.8249   &  89.5966  &  32.7
\\ \hline
\end{tabular}
\vspace{3mm}
\caption{Results of CPU experiments in \figureautorefname~\ref{fig:cpu_total}.}
\label{tab:cpu_total}
\vspace{-3mm}
\end{table}

\begin{table}[!h]
\centering
\begin{tabular}{|l||c|c|c|c|c|}
\hline
  Architecture & \# of params (M) & Gflops & GPU latency (ms) & CPU latency (ms) & mAP
\\ \hline \hline

PAFPN                 & 4.8606  &  7.3894  &  3.6284  &  80.6611  &  \textbf{33.3}
\\ \hline
LCPAN                 & 3.5535  &  6.6907  &  3.2626&  76.6075    &  31.9
\\ \hline
PAFPN (cat to sum)    & 4.7224  &  7.2399  &  3.6354&  79.3600    &  32.0
\\ \hline
LCAPN (cat to sum)    & \textbf{3.3845}  &  \textbf{6.3338}  & \textbf{3.2180}  &  \textbf{71.4919}    &  31.2
\\ \hline
SepFPN                & 4.3532  &  7.1149  &  3.5134  &  77.9030    &  32.3
\\ \hline
\end{tabular}
\vspace{3mm}
\caption{Results of FPN experiments in \figureautorefname~\ref{fig:fpn}.}
\label{tab:fpn}
\vspace{-3mm}
\end{table}

\clearpage

\subsection{Scale-up}

\begin{figure}[t]
  \centering
    \includegraphics[width=\columnwidth]{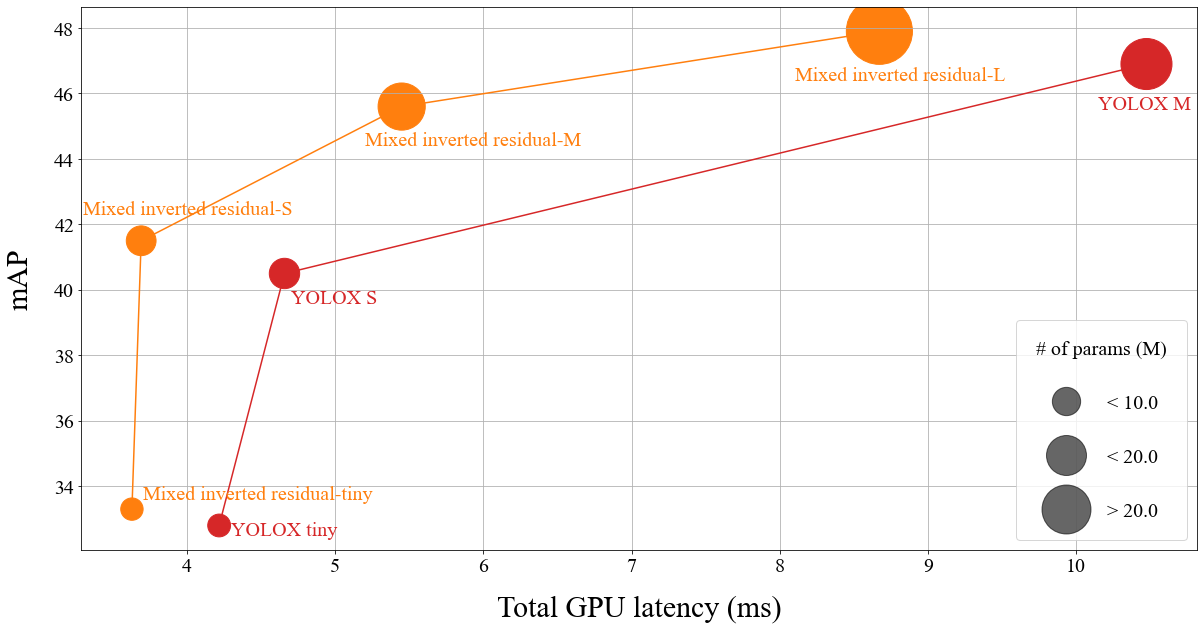}
  \caption{GPU target detectors at scale.}
  \label{fig:scale}
\end{figure}

In \sectionautorefname~\ref{sec:gpu}, we designed the GPU target detector which has a similar size to YOLOX tiny.
\figureautorefname~\ref{fig:scale} shows various sizes of YOLOX baseline and the GPU target detector searched in \sectionautorefname~\ref{sec:gpu} to verify the design policy establishsed on \sectionautorefname~\ref{sec:gpu} is efficient with other sizes.
Red dots denote YOLOX baselines of various sizes and orange dots denote GPU target detectors which are designed with the policy proposed in \sectionautorefname~\ref{sec:gpu}.
As shown in the figure, orange dots are better than red dots considering the latency-mAP trade-off.
As scaling up, improvements in latency are much better while mAP increases smoothly compared to YOLOX counterpart.
This trend will continue as the network grows in size.

\end{document}